\documentclass[10pt,twocolumn,letterpaper]{article}

\usepackage[pagenumbers]{cvpr} % To force page numbers, e.g. for an arXiv version

\usepackage{graphicx}
\usepackage{bm}
\usepackage{amsmath}
\usepackage{amssymb}
\usepackage{amsthm}
\usepackage{booktabs}

\usepackage{comment}
\usepackage{subcaption}
\usepackage{algorithm}
\usepackage{algorithmic}
\usepackage{multirow}
\usepackage{enumitem}
\usepackage[pagebackref,breaklinks,colorlinks]{hyperref}

\theoremstyle{definition}
\newtheorem{rmk}{Remark}
\newtheorem{theorem}{Theorem}

\usepackage[capitalize]{cleveref}
\crefname{section}{Sec.}{Secs.}
\Crefname{section}{Section}{Sections}
\Crefname{table}{Table}{Tables}
\crefname{table}{Tab.}{Tabs.}

\def\cvprPaperID{****} % *** Enter the CVPR Paper ID here
\def\confName{CVPR}
\def\confYear{2023}

\usepackage{marvosym}
\makeatletter
\def\@fnsymbol#1{\ensuremath{\ifcase#1\or \dagger\or \ddagger\or
		\mathsection\or \mathparagraph\or \|\or **\or \dagger\dagger
		\or \ddagger\ddagger \else\@ctrerr\fi}}
\makeatother

\begin{document}
	
\title{Multi-Agent Automated Machine Learning}
	
\author{
		Zhaozhi Wang\textsuperscript{$2,3$}\thanks{Work done during his Master program at Peking University}, Kefan Su\textsuperscript{$1$}, Jian Zhang\textsuperscript{$4$}, Huizhu Jia\textsuperscript{$1$}, Qixiang Ye\textsuperscript{$2,3$}, Xiaodong Xie\textsuperscript{$1$}, and Zongqing Lu\textsuperscript{$1$}\thanks{\Letter \ zongqing.lu@pku.edu.cn}
		\\
		\\
		\textsuperscript{$1$}Peking University \quad \textsuperscript{$2$}Peng Cheng Lab \quad \textsuperscript{$3$}University of Chinese Academy of Sciences  \quad \textsuperscript{$4$}Huawei
}

\maketitle
 
\begin{abstract}

%model the cooperation among modules in the machine learning (ML) pipeline observed in experiments and

In this paper, we propose \emph{multi-agent automated machine learning} (MA2ML) with the aim to effectively handle joint optimization of modules in automated machine learning (AutoML).
MA2ML takes each machine learning module, such as data augmentation (AUG), neural architecture search (NAS), or hyper-parameters (HPO), as an agent and the final performance as the reward, to formulate a multi-agent reinforcement learning problem.
%explore the ``cooperation" among modules
%
MA2ML explicitly assigns credit to each agent according to its marginal contribution to enhance cooperation among modules, and incorporates off-policy learning to improve search efficiency. 
Theoretically, MA2ML guarantees monotonic improvement of joint optimization.
%joint optimization from a perspective of monotonic policy improvement. 
%
Extensive experiments show that MA2ML yields the state-of-the-art top-1 accuracy on ImageNet under constraints of computational cost, \textit{e.g.}, $79.7\%/80.5\%$ with FLOPs fewer than 600M/800M. Extensive ablation studies verify the benefits of credit assignment and off-policy learning of MA2ML.

\end{abstract}

\section{Introduction}

Automated machine learning (AutoML) aims to find high-performance machine learning (ML) pipelines without human effort involvement. The main challenge of AutoML lies in finding optimal solutions in huge search spaces. 

\begin{figure}[t]
    \centering
    \includegraphics[width=.45\textwidth]{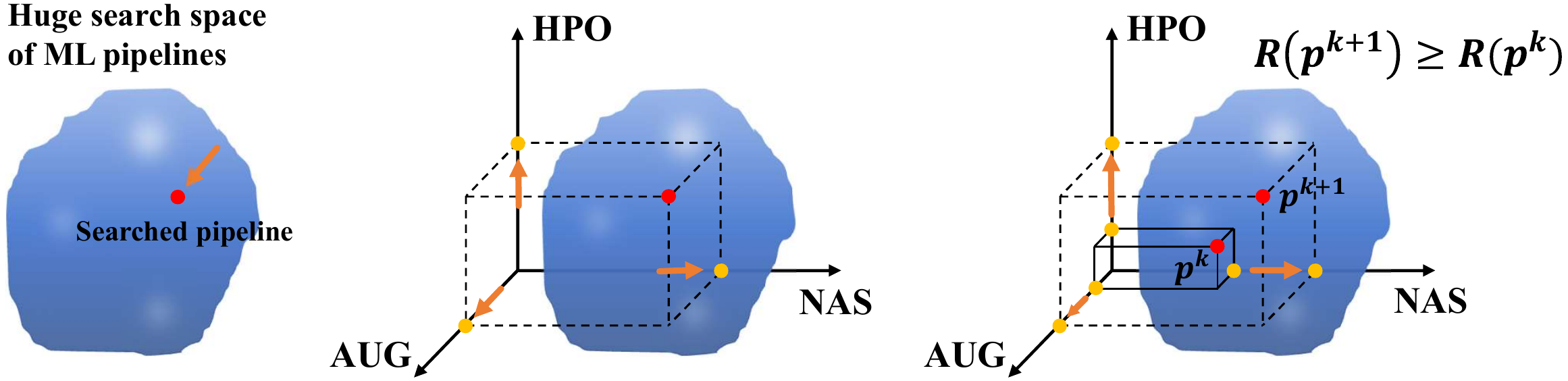}
    \vspace{-0.1cm}
    \caption{Search spaces of machine learning pipelines. \textit{Left}: single agent controls all modules, and the huge search space makes it ineffective to learn. \textit{Mid}: each agent controls one module, and the learning difficulty is reduced by introducing MA2ML. \textit{Right}: MA2ML guarantees monotonic improvement of the searched pipeline, where $p^k$ and $R(p^k)$ denote the $k$-th searched pipeline and its expected performance, respectively.}
    \label{fig: intro_fig}
    \vspace{-0.2cm}
\end{figure}

In recent years, reinforcement learning (RL) has been validated to be effective to optimize individual AutoML modules, such as data augmentation (AUG) \cite{cubuk2019autoaugment}, neural architecture search (NAS) \cite{zoph2016neural,zoph2018learning,mills2021l2nas}, and hyper-parameter optimization (HPO) \cite{2018Speeding}. 
However, when facing the huge search space (Figure \ref{fig: intro_fig} \textit{left}) and the joint optimization of these modules, the efficiency and performance challenges remain.

Through experiments, we observed that among AutoML modules there exists a cooperative relationship that facilities the joint optimization of modules. For example, a small network (ResNet-34) with specified data augmentation and optimized hyper-parameters significantly outperforms a large one (ResNet-50) with default training settings (76.8\% vs. 76.1\%).
In other words, good AUG and HPO alleviate the need for NAS to some extent.
Accordingly, we propose \textit{multi-agent automated machine learning} (MA2ML), which explores the cooperative relationship towards joint optimization of ML pipelines. 
In MA2ML, ML modules are defined as RL agents (Figure \ref{fig: intro_fig} \textit{mid}), which take actions to jointly maximize the reward, so that the training efficiency and test accuracy are significantly improved. 
Specially, we introduce credit assignment to differentiate the contribution of each module, such that all modules can be simultaneously updated. 
To handle both continuous (\textit{e.g.}, learning rate) and discrete (\textit{e.g.}, architecture) action spaces, MA2ML employs a multi-agent actor-critic method, where a centralized Q-function is learned to evaluate the joint action. Besides, to further improve search efficiency, MA2ML adopts off-policy learning to exploit historical samples for policy updates.

MA2ML is justified theoretically and experimentally. Theoretically, we prove that MA2ML guarantees monotonic policy improvement (Figure \ref{fig: intro_fig} \textit{right}), \textit{i.e.}, the performance of the searched pipeline monotonically improves in expectation. 
This enables MA2ML to fit the joint optimization problem and be adaptive to all modules in the ML pipeline, potentially achieving full automation. 
Experimentally, we take the combination of individual RL-based modules to form MA2ML-Lite, and compare their performance on ImageNet \cite{russakovsky2015imagenet} and CIFAR-10/100 \cite{krizhevsky2009learning} datasets. To better balance performance and computational cost, we add constraints of FLOPs in the experiment on ImageNet. Experiments show that MA2ML substantially outperforms MA2ML-Lite $w.r.t.$ both accuracy and sample efficiency, and MA2ML achieves remarkable accuracy compared with recent methods.

%Contributions
%\noindent 
{Our contributions are summarized as follows: }
\vspace{-2mm}
\begin{itemize}
\setlength\itemsep{0mm}
    \item  We propose MA2ML, which  utilizes credit assignment to differentiate the contributions of ML modules, providing a systematic solution for the joint optimization of AutoML modules.
    \item We prove the monotonic improvement of module policies, which enables to MA2ML fit the joint optimization problem and be adaptive to various modules in the ML pipeline.  
    \item MA2ML yields the state-of-the-art performance under constraints of computational cost, \textit{e.g.}, $79.7\%/80.5\%$ on ImageNet, with FLOPs fewer than 600M/800M, validating the superiority of the joint optimization of MA2ML. 
\end{itemize}

\section{Related work}

\noindent\textbf{Single-module Optimization.} AutoML has been applied to different modules in the ML pipeline, such as AUG, NAS, and HPO. For AUG, AutoAugment \cite{cubuk2019autoaugment} is an RL-based method to search augmentation policies. Fast AA \cite{lim2019fast} speeds up the search with less computing cost through Bayesian optimization. Faster AA \cite{hataya2020faster} is a differentiable method to optimize augmentation policy, which reduces computing cost further. PBA \cite{ho2019population} also reduces search cost by population-based training. Adversarial AA \cite{zhang2019adversarial} uses the idea of GAN \cite{goodfellow2014generative} to generate the hard augmentation policy for the model to improve robustness. AWS AA \cite{tian2020improving} improves AutoAugment by the implementation of augmentation-wise weight sharing. 

For NAS, there are three main types of methods, evolutionary methods \cite{real2017large,real2019regularized,yang2020cars}, RL-based methods \cite{zoph2016neural,zoph2018learning,Tan_2019_CVPR,mills2021l2nas,pham2018efficient}, and differentiable gradient-based methods \cite{liu2018darts,chu2020darts,he2020milenas,yang2020ista}. Evolutionary methods optimize the search through selection, recombination, and mutation in the neural architecture population. RL-based methods regard the neural architecture as a black box and the final accuracy as the reward. They typically use RL algorithms to solve the optimization problem. Differentiable gradient-based methods design a continuous representation for neural architecture space and make NAS a differentiable problem, which greatly improves search efficiency, compared with the other two types. 

For HPO, black-box methods \cite{bergstra2012random,hutter2011sequential} have been utilized for a long time. Meanwhile, some methods like \cite{kandasamy2017multi,wu2020practical} use multi-fidelity ways to accelerate optimization through the evaluation on proxy tasks. Recently, gradient-based methods \cite{maclaurin2015gradient,pedregosa2016hyperparameter,lorraine2020optimizing,shaban2019truncated} optimize hyper-parameters by calculating gradient respect to them, which reduces computing cost substantially. 

The above methods for single module optimization leave other modules of the ML pipeline fixed during search, either by expert knowledge or empirical setting, which may not be optimal when combining them. Joint optimization of multiple modules is a more plausible way to advance the ML pipeline.

\vspace{0.15cm}
\noindent\textbf{Joint Optimization.} Recent studies \cite{zela2018towards,klein2019tabular,dong2020autohas,dai2021fbnetv3} search for NAS and HPO jointly through RL or training an accuracy predictor, or \cite{kashima2021joint,wang2021daas} consider the joint optimization of AUG and NAS, and obtain convincing results by bi-level gradient-based optimization. DHA \cite{zhou2021dha} explores the joint optimization for AUG, NAS, and HPO through one-level optimization, which is achieved in a differentiable manner by optimizing a compressed lower dimensional feature space for NAS. However, the alternate optimization of the gradient-based method may get stuck at the non-stationary point with limited-order gradient descent.

Unlike existing joint optimization methods for AutoML, our MA2ML aims to update all modules, as well as guarantee the convergence of joint optimization.

\begin{figure*}[!t]
\centering
    \includegraphics[width=1\textwidth]{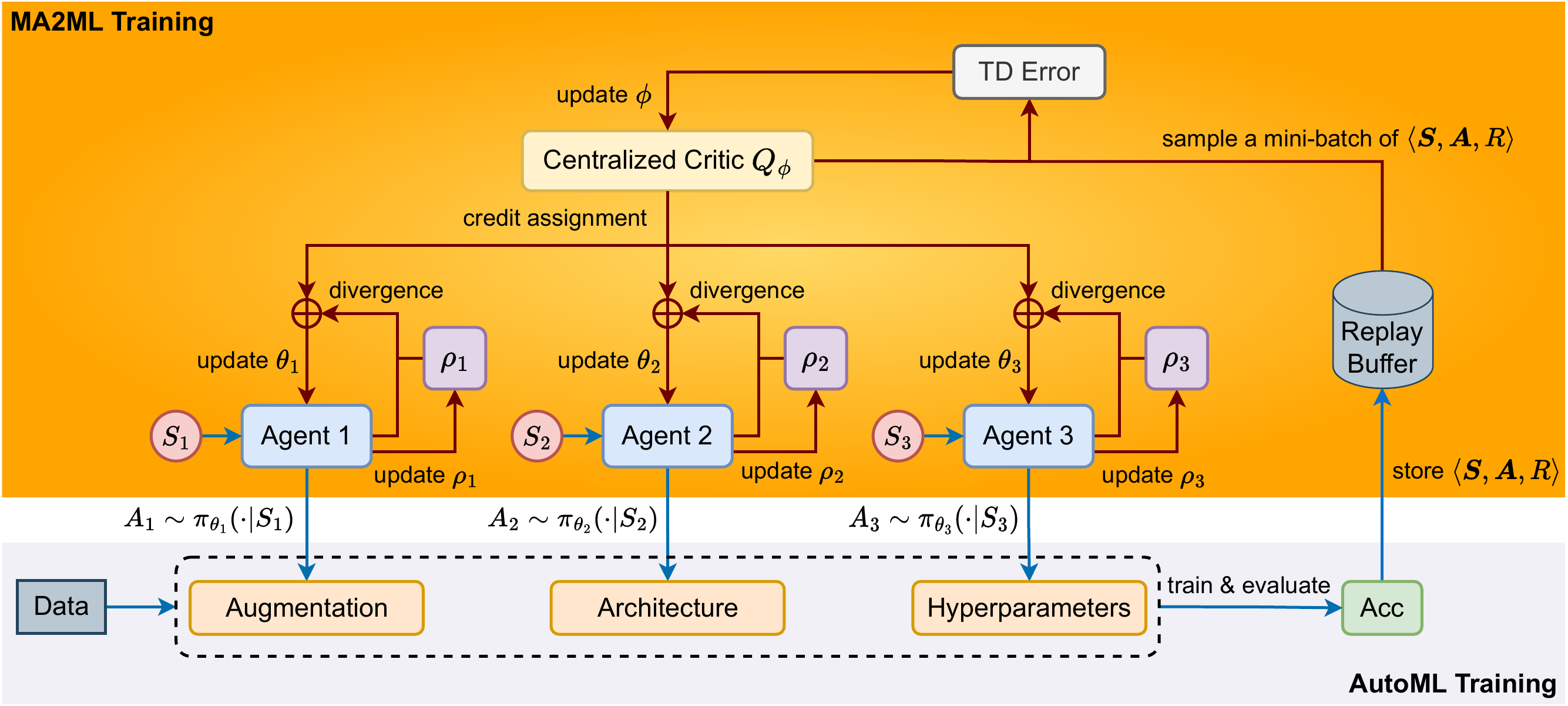}
    %\vspace*{-0.2cm}
    \caption{The framework of joint optimization with MA2ML. For AutoML training (lower panel), the ML pipeline is formed by the actions sampled from the policies of agents, then it is deployed for training on the dataset and to obtain the accuracy (reward). After that, the tuple $\langle \bm{S},\bm{A},R \rangle$ is stored in the replay buffer. For MA2ML training (upper panel), a mini-batch of $\langle \bm{S},\bm{A},R \rangle$ are sampled from the replay buffer to update the critic, policies, and target policies.}
    \label{fig: framework}
%\vspace*{-0.2cm}
\end{figure*}

\section{Method}

MA2ML is a general framework for joint optimization in AutoML and can be applied to any arbitrary combination of different modules. For ease of presentation, in the following, we use the joint optimization of AUG, NAS, and HPO for image classification tasks as an example to elaborate MA2ML.

\subsection{Action Space}
\label{searchspace}

Before formulating the joint optimization as an MARL problem, we first define an action space for AUG, NAS, and HPO. In AUG, we adopt the setting in AutoAugment \cite{cubuk2019autoaugment}. In NAS, we utilize the search space in NASNet \cite{zoph2018learning} and FBNetV3 \cite{dai2021fbnetv3} on different datasets to show that MA2ML is naturally agnostic to search spaces. In HPO, we design a search space for hyper-parameters, including learning rate, weight decay, etc.

For each module, we implement a controller to sample different choices from the action space sequentially. Note that MA2ML can be easily extended to support more modules in the ML pipeline, such as loss function search and mix-precision configuration search, by simply implementing the module by a controller and adding it to the framework.

\subsection{Joint Optimization}

Assuming there are $n$ modules for the AutoML pipeline, each module is modeled as an agent. For each agent $i$, it takes a random state $S_i$ as initial input and outputs the selected action $A_i$ in the action space sampled from its policy $\pi_i$, which is denoted as $A_i \sim \pi_{\theta_i}(\cdot | S_i)$ where the policy is parameterized by $\theta_i$. After all actions are determined, we train the network with the determined pipeline setting on the image classification task for epochs and evaluate top-1 accuracy on the validation set. We take the top-1 accuracy as the reward $R$ for all the agents. The objective is to maximize the expected reward $R$, represented by $J(\Theta)$:
\begin{equation}
\label{eq: goal}
    \begin{gathered}
        J(\Theta) = \mathbb{E}_{\bm{\pi}_{\Theta}(\bm{A}|\bm{S})}[R],
    \end{gathered}
\end{equation}
where $\bm{\pi}_{\Theta}(\bm{A}|\bm{S}) \triangleq \prod_{i=1}^n \pi_{\theta_i}(A_i|S_i)$ is the joint policy, and $\Theta$, $\bm{S}$ and $\bm{A}$ respectively denote the gather of $\theta_i$, $S_i$ and action $A_i$ of all agents (we may drop $\Theta$ or $\theta$ for brevity if there is no confusion). Consequently, we transform the joint optimization of the ML pipeline as an MARL problem, then we can rely on MARL methods to solve the problem.

\subsection{MA2ML-Lite}
\label{sec:baseline}
For the MARL problem defined in \eqref{eq: goal}, one method to solve it is to learn a policy individually for each module of AUG, NAS, and HPO. In the \texttt{AutoML} training phase, we sample an action according to the policy of each agent, and train the model according to the searched ML pipeline to calculate the reward $R$ (top-1 accuracy). In the \texttt{MA2ML} training phase, we use the reward $R$ to directly guide the update of each agent's policy. REINFORCE \cite{sutton2000policy} which can handle both discrete and continuous action, is used to calculate policy gradient, which is formulated as
\begin{equation}
\label{eq: baseline_method}
%    \begin{gathered}
        \nabla_{\theta_i} J(\Theta) = \mathbb{E}_{\bm{\pi}(\bm{A}|\bm{S})}[\nabla_{\theta_i} \log \pi_{\theta_i}(A_i|S_i)(R-b)],
%    \end{gathered}
\end{equation}
where $b$ is an exponential moving average of the previous rewards. The moving average is beneficial for reducing the variance of gradient estimate. Then, the policy of each agent $i$ is updated by gradient ascent using \eqref{eq: baseline_method}. Although $J(\Theta)$ depends on the joint policy of all agents, their policies are not \textit{really} jointly optimized by \eqref{eq: baseline_method}. In other words, for the obtained $R$, it is not able to tell the contribution of each agent, and thus cannot explicitly update their policies towards better ones. We term this RL method MA2ML-Lite, which is the lite version of MA2ML.

\subsection{Credit Assignment}

As AUG, NAS, and HPO jointly determine the final performance, if we use the reward $R$ to directly feedback to each agent as MA2ML-Lite does, each agent cannot distinctly determine whether the performed action is good or not. This poses a great challenge to the learning of policies.

To solve the challenge, we train a centralized critic to learn the action-value function (Q-function) and add a counterfactual baseline, inspired by \cite{foerster2018counterfactual}. The counterfactual baseline marginalizes out an agent’s action, while keeping other agents’ actions fixed. The difference between the value of taken actions and the counterfactual baseline measures the marginal contribution of each agent. The difference is used to update each agent's policy.

The centralized critic is denoted as $Q(\bm{S},\bm{A})$ or $Q(\bm{S},A_i,A_{-i})$, where $A_{-i}$ represents the joint action of all agents except agent $i$. The counterfactual baseline denoted as $b(\bm{S},A_{-i})$ for each agent $i$ is calculated as
\begin{equation}
\label{eq: baseline1}
%    \begin{gathered}
        b(\bm{S},A_{-i}) = \mathbb{E}_{A_i \sim \pi_{\theta_i}}[Q(\bm{S},A_i,A_{-i})].
%    \end{gathered}
\end{equation}
In AutoML, considering the large action space of different modules and to reduce the computation cost, we can use the sampled action $A_i$ as input, while keeping other actions fixed, to approximate the counterfactual baseline.

\subsection{Off-Policy Learning}

REINFORCE is an on-policy method, meaning that current policy can only be updated using the experiences obtained by itself. Considering it takes much time to train the model to receive top-1 accuracy, on-policy methods are very \textit{inefficient}. Thereby, we incorporate off-policy learning, such that the current policy can also use historical experiences generated during learning for updates, which helps a lot in improving search efficiency. 

To enable off-policy learning for actor-critic methods, a popular method is maximum-entropy RL, like SAC \cite{haarnoja2018soft}, which adds an entropy regularization in the objective as
\begin{equation}
\label{eq: goal_2}
    \begin{aligned}
        J(\Theta) 
        =& \mathbb{E}_{\bm{\pi}(\bm{A}|\bm{S})}[Q(\bm{S},\bm{A})-\lambda\log\bm{\pi}(\bm{A}|\bm{S})] \\
        =& \mathbb{E}_{\bm{\pi}(\bm{A}|\bm{S})}[Q(\bm{S},\bm{A})] +  \lambda\mathcal{H}(\bm{\pi}(\cdot|\bm{S})),
    \end{aligned}
\end{equation}
where $\mathcal{H}(\bm{\pi}(\cdot|\bm{S}))$ represents the entropy of $\bm{\pi}(\cdot|\bm{S})$, and $\lambda$ is the coefficient of entropy regularization. However, the entropy regularization biases the converged policy, \textit{i.e.}, the converged policy maximizes \eqref{eq: goal_2} instead of the original objective \eqref{eq: goal}.

To eliminate the bias of the converged policy as well as implement off-policy learning, inspired by \cite{su2021divergenceregularized}, we maintain a target policy $\rho_i$ for each agent, in addition to the original policy $\pi_i$, and add the divergence regularization between $\pi_i$ and $\rho_i$ in the reward, 
\begin{equation}
%\nonumber
\label{eq: goal_3}
    \begin{aligned}
        J(\Theta) 
        =& \mathbb{E}_{\bm{\pi}(\bm{A}|\bm{S})}[Q(\bm{S},\bm{A})-\lambda\log \frac{\bm{\pi}(\bm{A}|\bm{S})}{\bm{\rho}(\bm{A}|\bm{S})}] \\
        =& \mathbb{E}_{\bm{\pi}(\bm{A}|\bm{S})}[Q(\bm{S},\bm{A})] - \lambda D_{\mathrm{KL}}(\bm{\pi}(\cdot|\bm{S})\|\bm{\rho}(\cdot|\bm{S})),\\
    \end{aligned}
\end{equation}
where $\bm{\rho}\triangleq\prod_{i=1}^n \rho_i$ denotes the joint target policy and $D_{\mathrm{KL}}(\bm{\pi}(\cdot|\bm{S})\|\bm{\rho}(\cdot|\bm{S}))$ denotes the KL divergence between two distributions, $\bm{\pi}(\cdot|\bm{S})$ and $\bm{\rho}(\cdot|\bm{S})$. The divergence regularization is beneficial for exploration and stable policy improvement. More importantly, based on divergence policy iteration \cite{su2021divergenceregularized} we can further derive the theoretical result as follows. 

\begin{theorem}
\label{them1}
By iteratively applying divergence policy iteration and taking $\bm{\pi}^k$ as the joint target policy $\bm{\rho}^{k+1}$, the policy sequence of $\{\bm{\pi}^k\}$ converges and monotonically improves upon the \emph{original} optimization problem.

\begin{proof}
    Let $J_{\operatorname{init}}(\bm{\pi})$ be the original optimization objective and $J_{\operatorname{reg}}(\bm{\pi},\bm{\rho})$ be the optimization objective with divergence regularization given the fixed  target policy $\bm{\rho}$. Then they have the following definitions and relations.
    \begin{align}
        & J_{\operatorname{init}}(\bm{\pi}) = \mathbb{E}_{\bm{\pi}(\bm{A}|\bm{S})}[R], \\
        &J_{\operatorname{reg}}(\bm{\pi},\bm{\rho}) =\mathbb{E}_{\bm{\pi}(\bm{A}|\bm{S})}[R-\lambda\log \frac{\bm{\pi}(\bm{A}|\bm{S})}{\bm{\rho}(\bm{A}|\bm{S})}], \\
        & J_{\operatorname{reg}}(\bm{\pi},\bm{\rho}) = J_{\operatorname{init}}(\bm{\pi}) - \lambda D_{\operatorname{KL}}(\bm{\pi}(\cdot|\bm{S}) || \bm{\rho}(\cdot|\bm{S}) ).
        \label{reg}
    \end{align}
    
    Iteratively applying divergence policy iteration and taking $\bm{\pi}^k$ as the joint target policy $\bm{\rho}^{k+1}$ can be formulated as
    \begin{equation}
    \label{iter-eq}
        \bm{\pi}^{k+1} = \arg \max_{\bm{\pi}}  J_{\operatorname{reg}}(\bm{\pi},\bm{\rho}^{k+1}) = \arg \max_{\bm{\pi}}  J_{\operatorname{reg}}(\bm{\pi},\bm{\pi}^{k}). 
    \end{equation}
    Substituting \eqref{iter-eq} to \eqref{reg}, we have 
    \begin{equation}
    \begin{aligned}
        J_{\operatorname{init}}(\bm{\pi}^{k+1}) & \ge J_{\operatorname{init}}(\bm{\pi}^{k+1}) - \lambda D_{\operatorname{KL}}(\bm{\pi}^{k+1}(\cdot|\bm{S}) || \bm{\pi}^{k}(\cdot|\bm{S}) ) \\
        & = J_{\operatorname{reg}}(\bm{\pi}^{k+1},\bm{\pi}^{k}) \\
        & \ge J_{\operatorname{reg}}(\bm{\pi}^{k},\bm{\pi}^{k}) \\ 
        & = J_{\operatorname{init}}(\bm{\pi}^{k}). 
    \end{aligned}
    \label{iter_improve}
    \end{equation}
    The first inequality is from the non-negativity of the KL-divergence and the second inequality is from the definition of $\bm{\pi}^{k+1}$ in \eqref{iter-eq}.
    
    The conclusion $J_{\operatorname{init}}(\bm{\pi}^{k+1}) \ge J_{\operatorname{init}}(\bm{\pi}^{k}) $ (defined by \eqref{iter_improve}) means the original objective monotonically improves though we optimize the objective with divergence regularization.
    
    Moreover, the reward $R$ is bounded, so the original objective $J_{\operatorname{init}}(\bm{\pi})$ is bounded. The boundness and monotonic improvement of the sequence $J_{\operatorname{init}}(\bm{\pi}^{k})$ guarantees the convergence of the original objective sequence.
    %The sequence $J_{\operatorname{init}}(\bm{\pi}^{k})$ is bounded and improves monotonically which means this sequence will converge to sub-optimum.
\end{proof}
\end{theorem}

\begin{rmk}
Theorem~\ref{them1} guarantees the monotonic policy improvement and sub-optimum in the original optimization objective \eqref{eq: goal}. 
This is crucial for AutoML as we can control the search of MA2ML by $\lambda$, without deteriorating the optimality of the searched ML pipeline.
Therefore, Theorem~\ref{them1} lays the theoretical foundation of MA2ML. 
\end{rmk}

\subsection{Training}

In the AutoML training phase, we sample actions $\bm{A}$ from $\bm{\pi}$ to obtain the ML pipeline and train the model according to the pipeline to receive top-1 accuracy as the reward $R$. After that, we store the experience $\langle \bm{S}, \bm{A}, R\rangle$ in the replay buffer $\mathcal{D}$.

In the MA2ML training phase, we sample a mini-batch $\mathcal{B}$ of experiences from the replay buffer $\mathcal{D}$. We first update the centralized critic, parametrized by $\phi$, by gradient descent of the loss function $\mathcal{L}_Q$ as
\begin{equation}
\label{eq: critic_update}
    \begin{gathered}
    \mathcal{L}_Q = \mathbb{E}[(Q_\phi(\bm{S},\bm{A}) - R)^2], \\
    \phi = \phi - \eta_\phi \nabla_\phi \mathcal{L}_Q,
    \end{gathered}
\end{equation}
where $\eta_\phi$ denotes the learning rate of $\phi$. Then, we update the policy and the target policy for each agent. The counterfactual baseline with divergence regularization is modified as
\begin{equation}
%\nonumber
\label{eq: baseline}
    \begin{array}{ll}
        b(\bm{S},A_{-i}) = \mathbb{E}_{A_i \sim \pi_i}[Q(\bm{S},A_i,A_{-i}) - \lambda \log \frac{\bm{\pi}(\bm{A}|\bm{S})}{\bm{\rho}(\bm{A}|\bm{S})}].
    \end{array}
\end{equation}
The gradient for the policy of each agent is calculated and the policy is updated by gradient ascent as
\begin{equation}
\begin{gathered}
\label{eq: actor_update}
    \begin{aligned}
        \nabla_{\theta_i}\mathcal{L}_{\pi_{\theta_i}} & = \mathbb{E}[\nabla_{\theta_i}\log\pi_{\theta_i}(A_i|S_i)( Q(\bm{S},\bm{A}) \\
        & \qquad -\lambda\log\frac{\pi_{\theta_i}(A_i|S_i)}{\rho_i(A_i|S_i)}-\mathbb{E}_{A_i\sim\pi_i}[Q(\bm{S},\bm{A})] \\
        & \qquad\qquad+\lambda D_{\mathrm{KL}}(\pi_i(\cdot|S_i)\| \rho_i(\cdot|S_i)))],
    \end{aligned}\\
    \theta_i = \theta_i + \eta_\theta \nabla_{\theta_i} \mathcal{L}_{\pi_{\theta_i}},
\end{gathered}
\end{equation}
where $\eta_\theta$ is the learning rate of the policy. The target policy of each agent is parameterized by $\bar{\theta}_i$, and is updated as 
\begin{equation}
\label{eq: target_policy_update}
    \begin{gathered}
        \bar{\theta}_i = (1-\tau)\bar{\theta}_i + \tau\theta_i,
    \end{gathered}
\end{equation}
where $\tau$ is an empirically determined parameter.

For completeness, the framework of MA2ML is depicted in Figure \ref{fig: framework} and the learning algorithm of MA2ML is summarized in Algorithm~\ref{alg: MA2ML}.

\begin{algorithm}[!t]
\caption{MA2ML}
\label{alg: MA2ML}
\begin{algorithmic}[1]
\STATE initialize the critic $\phi$, and the policy $\theta_i$ and target policy $\bar{\theta}_i$ for each agent $i$ 
\FOR{$iter=1$ to $max\_iter$}
\STATE receive initial state $S_i$ for each agent $i$
\STATE sample $A_i \sim \pi_{\theta_i}(\cdot|S_i)$ for each agent $i$
\STATE determine the pipeline setting according to $\bm{A}$
\STATE train the pipeline for a number of epochs on dataset
\STATE obtain the top-1 accuracy on the validation set as $R$
\STATE store $\langle \bm{S},\bm{A},R \rangle$ in the replay buffer $\mathcal{D}$
\STATE sample a mini-batch $\mathcal{B}$ from $\mathcal{D}$
\STATE update the critic $\phi$ by \eqref{eq: critic_update}
\STATE update each agent's policy $\theta_i$ by \eqref{eq: actor_update}
\STATE update each agent's target policy $\overline{\theta}_i$ by \eqref{eq: target_policy_update}
\ENDFOR
\STATE choose top-k pipelines in terms of top-1 accuracy
\STATE retrain the pipelines till convergence
\STATE obtain the highest accuracy on the test set
\end{algorithmic}
\end{algorithm}

\section{Experiments}

Image classification experiments of MA2ML and MA2ML-Lite are performed on ImageNet while ablation studies are carried on CIFAR-10/100\footnote{Detailed experiment settings, results, and the search cost are available in Appendix.}. 

\begin{figure}[!t]
    \centering
    \includegraphics[width=0.4\textwidth]{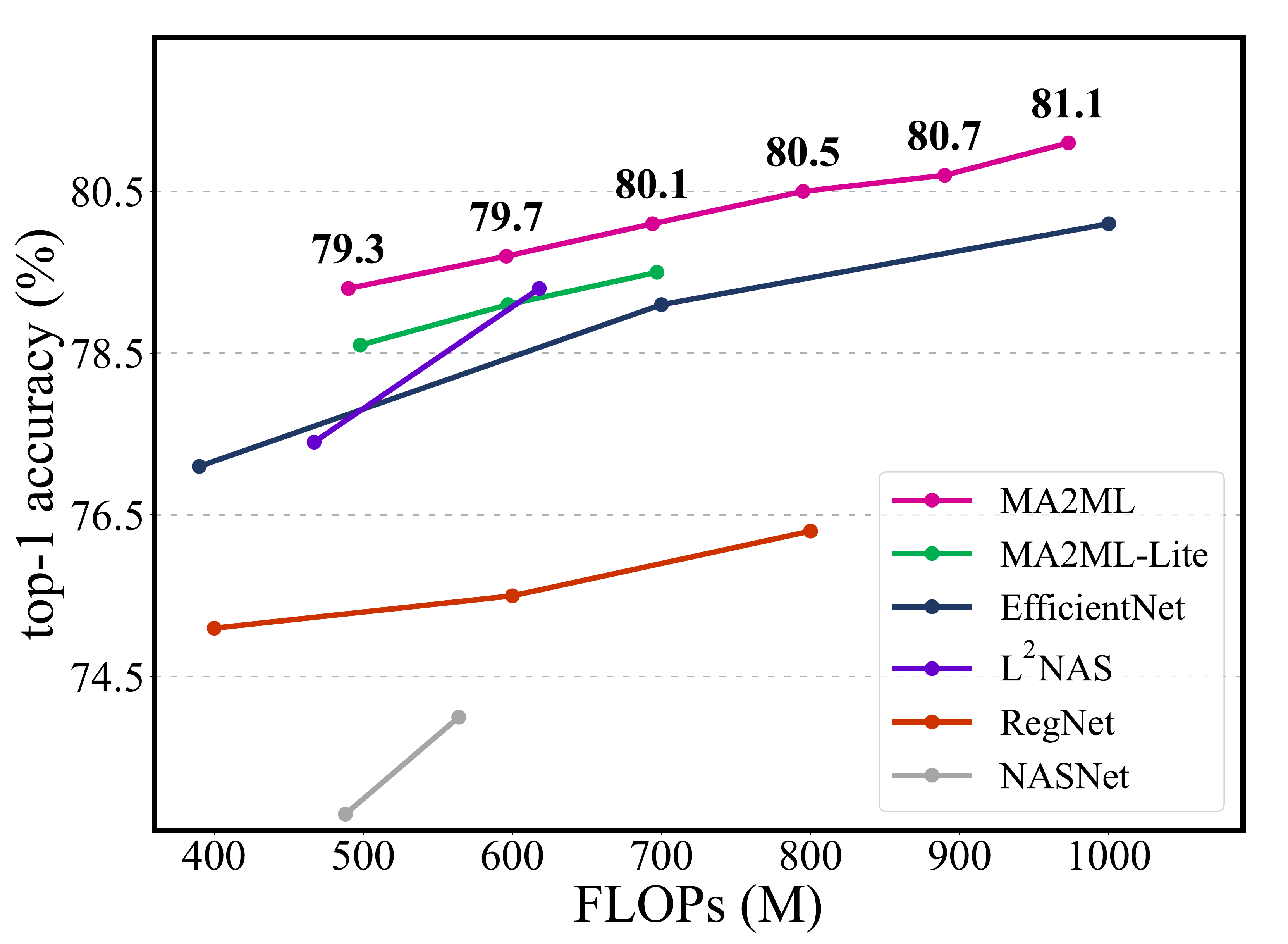}
 %   \vspace*{-0.2cm}
    \caption{Comparison of MA2ML with other AutoML methods on ImageNet.}
    \label{fig: acc_flops_comparison}
 %   \vspace*{-0.2cm}
\end{figure}

\subsection{ImageNet}

\noindent\textbf{Setting.} Considering the large search cost, we adopt the setting in FBNetV3 \cite{dai2021fbnetv3}, $i.e.$, randomly sampling 200 classes from the entire dataset as the training set and taking out 10K images from it to form a validation set. The sub-set with 200 classes is termed ImageNet-200.

The action space mentioned in Section \ref{searchspace} is employed for the experiment on ImageNet. For AUG, each pipeline is configured with an augmentation policy, which consists of 25 sub-policies. Each sub-policy contains two augmentation operations, which are determined by three dimensions including operation type, probability, and magnitude. Operation type has 15 options, probability ranges from $0.0$ to $1.0$ with step $0.1$, and magnitude consists of 10 levels.

For NAS and HPO, we use the search space in FBNetV3. We search for input resolution, kernel size, expansion, number of channels per layer, and depth as architecture configuration, which is based on the inverted residual block~\cite{inverted_residuals}. For HPO, we search for the optimizer type, initial learning rate, weight decay, mixup ratio \cite{zhang2018mixup}, drop out ratio, stochastic depth drop ratio \cite{huang2016deep}, and whether to use exponential moving average (EMA) \cite{kingma2015adam}. Overall, AUG, NAS, and HPO contain $10^{160}$, $10^{17}$, $10^{7}$ candidates, respectively.

\begin{table*}[!t]
\renewcommand\arraystretch{1}
  \begin{center}
    \caption{Top-1 accuracy (\%) and FLOPs of state-of-the-art AutoML methods on ImageNet, where NARS denotes neural architecture-recipe search. All compared models have computational cost close to 600M FLOPs for a fair comparison.}
    \vspace*{-0.2cm}
    \label{table: imagenet}
    \begin{tabular}{lllll}
      \toprule
      \textbf{Model} & \textbf{Acc (\%)} & \textbf{FLOPs (M)} & \textbf{Method} & \textbf{Search Modules}\\ \midrule
      DARTS \cite{liu2018darts} & 73.3 & 574 & gradient & NAS\\
      NASNet \cite{zoph2018learning} & 74.0 & 564 & RL & NAS\\
      MiLeNAS \cite{he2020milenas} & 75.3 & 584 & gradient & NAS\\
      RMI-NAS \cite{Zheng_2022_CVPR} & 75.3 & 657 & Random Forest & NAS\\
      RegNetY \cite{radosavovic2020designing} & 75.5 & 600 & pop. param.$^{*}$ & NAS\\
      ROME \cite{wang2020rome} & 75.5 & 556 & gradient & NAS\\
      AmoebaNet-C \cite{real2019regularized} & 75.7 & 570 & evolution & NAS\\
      PC-DARTS \cite{xu2020pcdarts} & 75.8 & 597 & gradient & NAS\\
      BaLeNAS \cite{Zhang_2022_CVPR} & 75.8 & 597 & gradient & NAS\\
      ISTA-NAS \cite{yang2020ista} & 76.0 & 638 & gradient & NAS\\
      Shapley-NAS \cite{Xiao_2022_CVPR} & 76.1 & 582 & gradient & NAS\\
      DAAS \cite{wang2021daas} & 76.6 & 698 & gradient & AUG+NAS\\
      DHA \cite{zhou2021dha} & 77.4 & - & gradient & AUG+NAS+HPO\\
      MIGO-NAS \cite{migo-nas} & 78.3 & 595 & MIGO & NAS\\
      OFA$^{\dagger}$ \cite{cai2020once} & 79.0 & 595 & gradient & NAS\\
      EfficientNet-B1 \cite{tan2019efficientnet} & 79.1 & 700 & RL & NAS\\
      FBNetV3$^{\ddagger}$ \cite{dai2021fbnetv3} & 79.2 & 550 & NARS & NAS+HPO\\
      L$^{2}$NAS \cite{mills2021l2nas} & 79.3 & 618 & RL & NAS\\ \midrule
      \textbf{MA2ML-A} & 79.3 & 490 & MARL & AUG+NAS+HPO \\
      \textbf{MA2ML-B} & \textbf{79.7} & 596 & MARL & AUG+NAS+HPO \\\bottomrule
      \multicolumn{5}{l}{${^*}${\footnotesize Population parameterization.} $^{\dagger}${\footnotesize Results are given in \cite{mills2021l2nas} without distillation.}}\\
      %\multicolumn{5}{l}{$^{\dagger}${\footnotesize Results are given in \cite{mills2021l2nas} without distillation.}}\\
      \multicolumn{5}{l}{$^{\ddagger}${\footnotesize Results are reproduced according to 600M FLOPs constraint without distillation.}}
    \end{tabular}
  \end{center}
\vspace*{-0.2cm}
\end{table*}

\begin{table}[!t]
\renewcommand\arraystretch{1}
\setlength{\tabcolsep}{6pt}
    \begin{center}
    \caption{Comparison of MA2ML-A/B/C with MA2ML-Lite-A/B/C on ImageNet.}
    \vspace{-0.2cm}
    \label{table: comparison_with_Lite}
    \begin{tabular}{lcc}
        \toprule
        \multirow{2}{*}{\textbf{Model}} & \multicolumn{2}{c}{\textbf{A/B/C}}\\ \cmidrule{2-3}
        & \textbf{Acc (\%)} & \textbf{FLOPs (M)}\\ \midrule
        MA2ML-Lite  & 78.6 / 79.1 / 79.5 & 498 / 597 / 697 \\
        \textbf{MA2ML} & 79.3 / 79.7 / 80.1 & 490 / 596 / 694 \\
        \bottomrule
    \end{tabular}
    \end{center}
\vspace{-0.2cm}
\end{table}

To better balance the performance and computational cost of the searched AutoML pipeline,  constraints of FLOPs are added. We adopt the multi-objective reward function $R=\operatorname{Acc}(m)\times\left[\frac{\operatorname{FLOPs}(m)}{\operatorname{FLOPs\_constraint}}\right]^w$ in \cite{tan2019mnasnet}, where $\operatorname{Acc}(m)$ and $\operatorname{FLOPs}(m)$ respectively denote top-1 accuracy and floating point operations of the model, $\operatorname{FLOPs\_constraint}$ represents the constraint of FLOPs, and $w$ is used to adjust the tradeoff between accuracy and FLOPs. We set $w=-0.07$ and $\operatorname{FLOPs\_constraint}=$ 600M/900M for two trials using MA2ML. Besides, the MA2ML-Lite runs $\operatorname{FLOPs\_constraint}=$ 600M and compared with MA2ML.

We generate 24 ML pipelines for each batch, and train each model for 100 epochs on ImageNet-200 to receive top-1 accuracy. For one trial, 1992 pipelines are searched on ImageNet-200 in total. After the search, 20 pipelines of the highest accuracy are retrained under the given constraint on the entire ImageNet dataset for 400 epochs. The pipeline of the highest accuracy is selected as the search result. For the search results obtained by MA2ML and MA2ML-Lite with $\operatorname{FLOPs\_constraint}=$ 600M, we retrain those with FLOPs fewer than 500M/600M/700M and pick the best as MA2ML-A/B/C and MA2ML-Lite-A/B/C. For the search results obtained with $\operatorname{FLOPs\_constraint}=$ 900M, we retrain those with FLOPs fewer than 800M/900M/1G and pick the best as MA2ML-D/E/F.

\vspace{0.15cm}
\noindent\textbf{Performance.} On ImageNet, MA2ML-A/B/C/D/E/F achieves $79.3\%$, $79.7\%$, $80.1\%$, $80.5\%$, $80.7\%$, and $81.1\%$ top-1 accuracy. Figure \ref{fig: acc_flops_comparison} gives the accuracy under different FLOPs of MA2ML and other methods. Since we do not use distillation in our experiments, results achieved by implementing distillation are not drawn here. The curve of MA2ML is plotted with the results of MA2ML-A/B/C/D/E/F, which is the highest among all the compared methods with a large margin. Figure \ref{fig: acc_flops_comparison} also indicates that MA2ML can always perform better with larger constraints of computation cost.

%Considering most state-of-the-art methods provide results with FLOPs close to 600M on ImageNet, 

In Table \ref{table: imagenet}, we compared MA2ML-A, MA2ML-B, and other methods with similar FLOPs. MA2ML-A outperforms all listed AutoML methods with only 490M FLOPs (L$^{2}$NAS has the same performance but with much larger FLOPs), and MA2ML-B with 596M FLOPs obtains the performance with large improvement compared with the results of other methods close to 600M FLOPs.

In Table \ref{table: comparison_with_Lite}, MA2ML-Lite-A/B/C achieves $78.6\%$, $79.1\%$ and $79.5\%$. The comparison shows that MA2ML outperforms MA2ML-Lite by $0.6\%$, which is \textbf{\textit{attributed to credit assignment and off-policy learning of MA2ML}}.

\begin{figure}[t]
    \centering
    \includegraphics[width=0.23\textwidth]{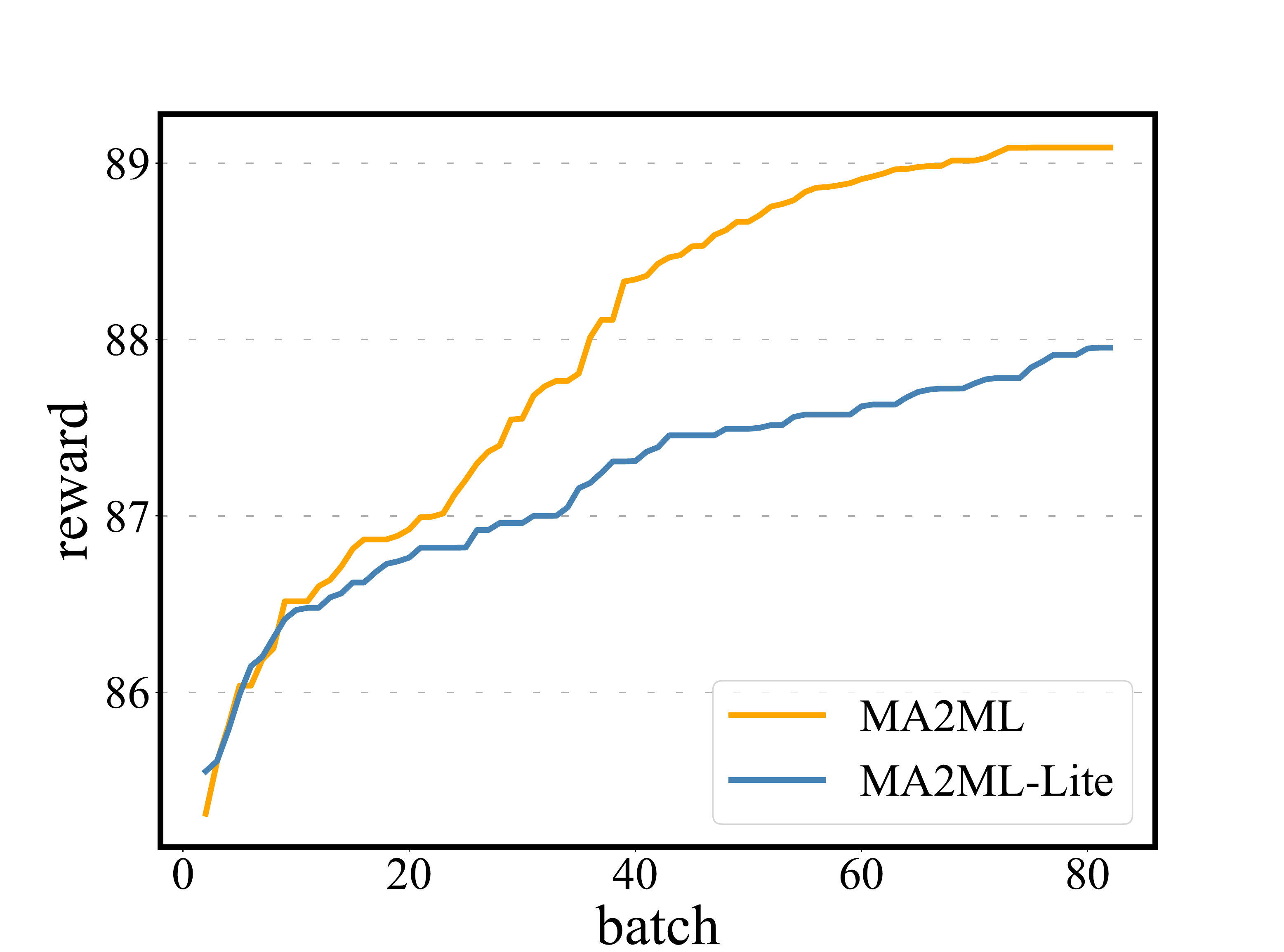}
    \includegraphics[width=0.23\textwidth]{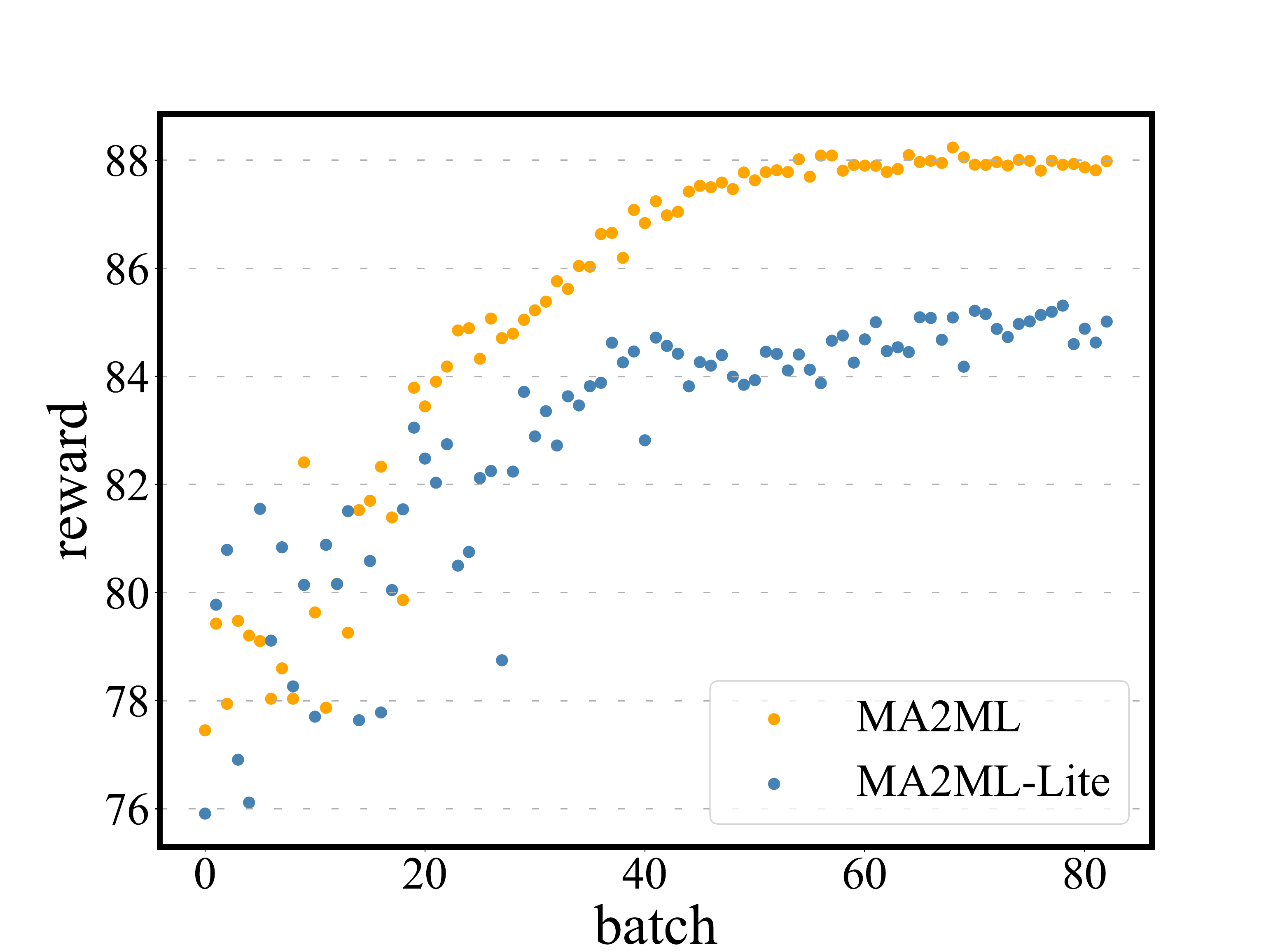}
    \caption{Learning patterns of MA2ML and MA2ML-Lite on ImageNet-200 ($\operatorname{FLOPs\_constraint}=$ 600M). \textit{Left}: the average reward curves of top-20 pipelines in terms of patch numbers. \textit{Right}: the scatter plot for average rewards of pipelines in each batch.}
    \label{fig: imagenet200}
\end{figure}

Figure \ref{fig: imagenet200} presents the average reward curves of top-20 pipelines and the scatter plot of average reward in each batch of MA2ML and MA2MLs-Lite on ImageNet-200 ($\operatorname{FLOPs\_constraint}=$ 600M). 
The curves illustrate that the superiority of MA2ML at the early stage is not significant, but surpasses MA2ML-Lite a lot in the mid-term of the search and keeps the lead till the end.  
The scatter depicts the batch-level average reward, where MA2ML also greatly outperforms MA2ML-Lite from the mid-term.
Moreover, the batch-level average reward of MA2ML is also more stable than MA2ML-Lite, which demonstrates \textbf{\textit{the benefit of monotonic policy improvement}}. 

MA2ML takes almost the same search cost as MA2ML-Lite and other RL-based methods, because the major part of search cost is the training time to receive top-1 accuracy and the cost of MARL in MA2ML is negligible. 
So the performance gain of MA2ML over MA2ML-Lite takes no more computation cost. The performance on ImageNet proves the effectiveness and generality of MA2ML, and jointly searching for ML modules is necessary.

\subsection{CIFAR}
\label{cifar_exp}

\noindent\textbf{Setting.} We also use the action space mentioned in Section \ref{searchspace} on CIFAR-10 and CIFAR-100. For AUG, we generate 25 augmentation sub-policies, which is the only difference from AUG module on ImageNet. For NAS, we utilize the search space in NASNet, and search for a normal cell and a reduction cell. We construct the whole network with 17 cells, and the 5th and 11th are designed as reduction cells. A cell is composed of 5 sequential blocks. To create a new block, we search for two existing blocks, choose two operations and apply them to the selected blocks, and select a way to combine the outputs of the last step. The operation space consists of 13 options and the combination contains element-wise addition and concatenation. For HPO, we choose SGD as the optimizer, fix the batch size, and search for warmup learning rate and weight decay, since these two hyper-parameters affect much the performance by experience. We discretize the search space of them into 4 concrete values and use 10 steps to determine them on average. Overall, AUG, NAS and HPO space contains $10^{32}$, $10^{37}$, $10^{12}$ candidates, respectively.

On CIFAR-10 and CIFAR-100, we sample 24 ML pipelines for each batch according to the policies of agents, and train the model using each pipeline for 12 epochs to receive top-1 accuracy as the reward. We have in total 5016 pipelines on CIFAR-10 and 4008 on CIFAR-100 and choose the top 30 pipelines in terms of top-1 accuracy to train for 600 epochs. 
As the parameter size is not constrained on CIFAR-10 and CIFAR-100, the searched models tend to be larger than those of the compared methods.
The ML pipeline of the highest top-1 accuracy on the test set is selected to train the final models. 
The pipeline is trained for 5 trials for both MA2ML and MA2ML-Lite. The mean and standard deviation are reported as the performance.

\begin{table*}[ht]
\renewcommand\arraystretch{1}
\setlength{\tabcolsep}{4.5pt}
  \begin{center}
    \caption{
    Top-1 accuracy (\%) and parameter size (MB) of compared AutoML methods on CIFAR-10 and CIFAR-100. The results of MA2ML and MA2ML-Lite are mean and standard deviation of 5 trials of the best ML pipeline.}
    \label{table: cifar}
    \vspace*{-2mm}
    \begin{tabular}{llllll}
      \toprule
      \multirow{2}{*}{\textbf{Model}} & \multicolumn{2}{c}{\textbf{CIFAR-10}} & \multicolumn{2}{c}{\textbf{CIFAR-100}} & \multirow{2}{*}{\textbf{Method}} \\ \cmidrule{2-5}
      & \textbf{Acc (\%)} & \textbf{Param (M)} & \textbf{Acc (\%)} & \textbf{Param (M)} & \\ \hline
      NASNet-A \cite{zoph2018learning} & $97.60$ & $27.6$ & - & - & RL \\
      ENAS \cite{pham2018efficient} & $97.11$ & $4.6$ & $80.57$ & $4.6$ & RL \\
      L$^{2}$NAS \cite{mills2021l2nas} & $97.51 {\scriptstyle \pm 0.12}$ & $3.8$ & $82.24{\scriptstyle \pm0.19}$ & $3.5$ & RL \\ 
      AmoebaNet \cite{real2019regularized} & $97.45 {\scriptstyle \pm 0.05}$ & $2.8$ & $81.07$ & $3.1$ & evolution \\
      DARTS \cite{liu2018darts} & $97.24 {\scriptstyle \pm 0.09}$ & $3.3$ & $82.64 {\scriptstyle \pm0.44}$ & $3.3$ & gradient \\
      DARTS- \cite{chu2020darts} & $97.41 {\scriptstyle \pm0.08}$ & $3.5$ & $82.49 {\scriptstyle \pm0.25}$ & $3.3$ & gradient \\
      MiLeNAS \cite{he2020milenas} & $97.49 {\scriptstyle \pm0.11}$ & $3.9$ & - & - & gradient \\
      ISTA-NAS \cite{yang2020ista} & $97.64 {\scriptstyle \pm0.06}$ & $3.4$ & $83.10 {\scriptstyle \pm0.11}$ & - & gradient \\
      \midrule
      AutoHAS \cite{dong2020autohas} & $95.0$ & - & $78.4$ & - & RL \\
      Joint Search \cite{kashima2021joint} & $97.46 {\scriptstyle \pm0.09}$ & - & $83.81 {\scriptstyle \pm0.49}$ & - & gradient \\
      DAAS \cite{wang2021daas} & $97.76 {\scriptstyle \pm0.10}$ & $4.0$ & $84.63 {\scriptstyle \pm0.31}$ & $3.8$ & gradient \\
      DHA \cite{zhou2021dha} & $98.11 {\scriptstyle \pm0.26}$ & - & $83.93 {\scriptstyle \pm0.23}$ & - & gradient \\
      \midrule
      MA2ML-Lite & $97.70 {\scriptstyle \pm 0.10}$ & $7.8$ & $84.80 {\scriptstyle \pm0.12}$ & 9.0 & MARL \\
      \textbf{MA2ML} & $97.77 {\scriptstyle \pm0.07}$ & $9.0$ & $85.08 {\scriptstyle \pm0.14}$ & $7.7$ & MARL \\ \bottomrule
    \end{tabular}
  \end{center}
\vspace*{-2mm}
\end{table*}

\vspace{0.15cm}
\noindent\textbf{Results on CIFAR-10/100.} MA2ML respectively achieves $97.77\pm0.07\%$ and $85.08\pm0.14\%$ top-1 accuracies on CIFAR-10 and CIFAR-100, higher than those ($97.70\pm0.10\%$ and $84.80\pm0.12\%$) of MA2ML-Lite. Table \ref{table: cifar} shows the results achieved by MA2ML, MA2ML-Lite, and other outstanding AutoML methods. Compared with the state-of-the-art AutoML methods, MA2ML achieves very competitive performance.

\begin{figure}[ht]
    \centering
    \includegraphics[width=0.23\textwidth]{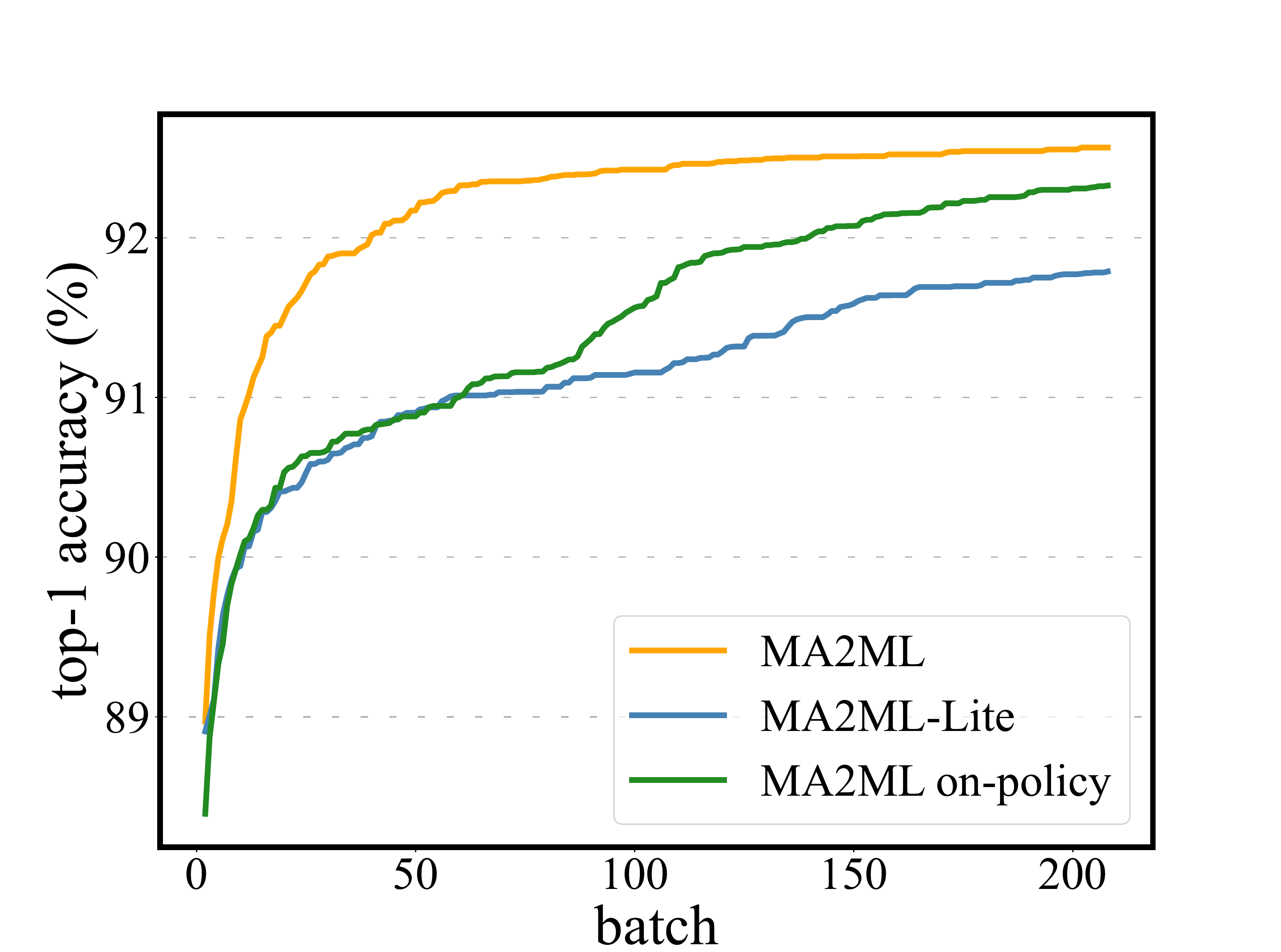}
    \includegraphics[width=0.23\textwidth]{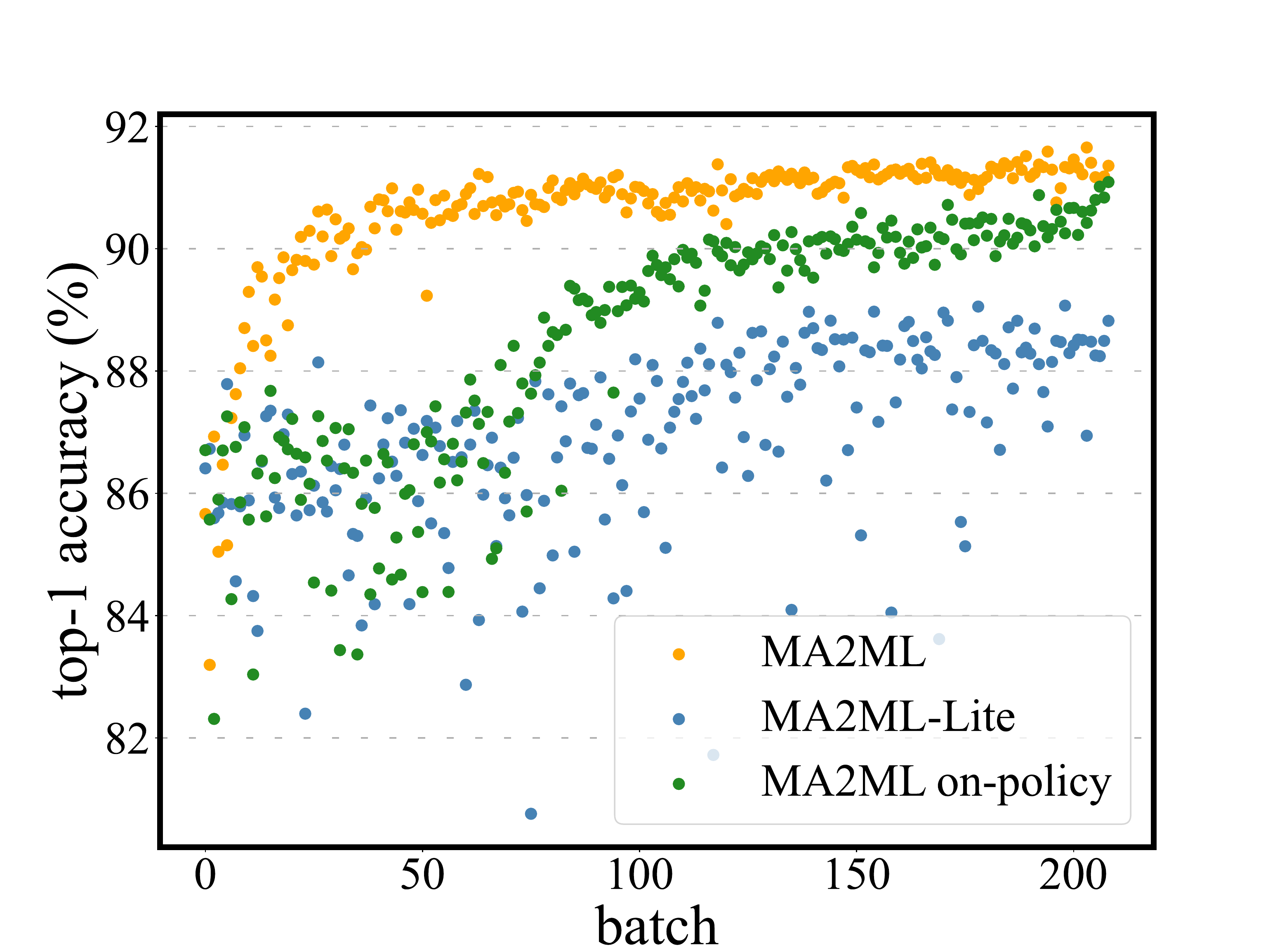}
    \caption{Learning patterns of MA2ML, MA2ML on-policy, and MA2ML-Lite on CIFAR-10. \textit{Left}: average accuracy curve of top 30 pipelines in terms of the number of batches. \textit{Right}: the scatter plot for the average accuracy of different pipelines in each batch. MA2ML outperforms MA2ML on-policy and MA2ML-Lite consistently in terms of accuracy and sample efficiency during training.}
    \label{fig: cifar10}
\end{figure}

\vspace{0.15cm}
\noindent\textbf{Ablation on off-policy learning and credit assignment.} To show search efficiency improved by off-policy learning, we implement MA2ML on-policy version as an ablation on CIFAR-10. Figure \ref{fig: cifar10} illustrates the average accuracy curves of top-30 pipelines and the scatter plot for average accuracy in each batch of MA2ML, MA2ML-Lite, and MA2ML on-policy during learning on CIFAR-10. 
The performance of MA2ML surpasses MA2ML-Lite at early stages and keeps the lead all the way, which indicates the effectiveness and high sample efficiency of MA2ML on CIFAR-10. From the scatter diagram, one can see that MA2ML outperforms MA2ML-Lite by a large margin from the start. Compared with MA2ML on-policy, MA2ML surpasses it a lot at the beginning of the search and leads all the way, which validates the search efficiency gains caused by off-policy learning. Besides, the performance gap between the MA2ML on-policy and MA2ML-Lite in the late search stages indicates the superiority of credit assignment\footnote{The learning pattern of MA2ML and MA2ML-Lite on CIFAR-100 is available in Appendix, which further validates MA2ML's superiority.}.

\begin{table}[h]
\renewcommand\arraystretch{1}
\setlength{\tabcolsep}{2pt}
    \begin{center}
    \caption{Comparison of NAS+AUG+HPO, AUG+HPO, and HPO on CIFAR-100.}
    \label{table: comparison_of_sub_module}
    \vspace{-2mm}
    %\begin{small}
    \begin{tabular}{ccccc}
        \toprule
        {\textbf{Module}} & {\textbf{Baseline}} & {\textbf{HPO}} & {\textbf{AUG+HPO}} & {\textbf{NAS+AUG+HPO}}\\\midrule
        Acc(\%)  & 73.50 & 74.39 & 77.37 & 85.08\\
        \bottomrule
    \end{tabular}
    %\end{small}
    \end{center}
\vspace{-2mm}
\end{table}

\vspace{0.15cm}
\noindent\textbf{Ablation on ML Modules.}  On CIFAR-100, the ablation studies for ML modules are performed. We fix the network architecture as ResNet-56 and run the optimization of HPO and AUG+HPO. In Table \ref{table: comparison_of_sub_module}, one can see that HPO improves 0.89\% accuracy (74.39\% vs. 73.50\%) over the baseline (ResNet-56 using default hyper-parameters) while AUG and NAS respectively improve 2.98\% and 7.71\%, which are all significant margins.

\section{Conclusion}

We proposed MA2ML, a general framework for the joint optimization of ML pipelines. MA2ML transforms the joint optimization of modules as an MARL problem, and provides a method with theoretical guarantee. Empirically, we investigated the joint optimization of AUG, NAS, and HPO for image classification tasks on ImageNet and CIFAR-10/100. MA2ML yields state-of-the-art top-1 accuracy on ImageNet under constraints of computational cost with similar settings, \emph{e.g.}, $79.7\%/80.5\%$ with FLOPs fewer than 600M/800M. MA2ML substantially outperforms MA2ML-Lite and MA2ML on-policy, which empirically verified the benefit of credit assignment and off-policy learning of MA2ML.

MA2ML has several advantages over other AutoML methods. First, MA2ML can be applied to the joint optimization of any arbitrary combination of different ML modules. Second, MA2ML is agnostic to search spaces, so we can choose suitable search spaces for different datasets. Third, MA2ML can also be easily generalized to different tasks by taking the evaluation (e.g., mIOU and pixel accuracy for semantic segmentation, and mAP for object detection) as the reward. Fourth, MA2ML is an RL-based method, which optimizes reward instead of training loss. Consequently, the incompatibility of different ML modules that may exist in the methods based on training loss, \emph{e.g.}, gradient-based methods, is naturally avoided. Last but not the least, the search of MA2ML can be easily controlled without affecting the optimality of the performance.

MA2ML may have some limitations. From the experiments, we can find that there exists a gap between the rank of short-term and final long-term training accuracy, which causes the superiority of MA2ML’s final performance over MA2ML-Lite not as large as in the search. Moreover, although MA2ML provides theoretical guarantee of the performance, as an MARL-based method, the search cost is large as other RL-based AutoML methods, since receiving the final reward costs much. How to address these two limitations is left as future work.

{\small
\bibliographystyle{ieee_fullname}
\bibliography{egbib_CVPR2023}
}

\clearpage

\appendix

\section{MA2ML and MA2ML-Lite on CIFAR-100}

\begin{figure}[h]
%\vspace{-4mm}
    \centering
    \includegraphics[width=0.23\textwidth]{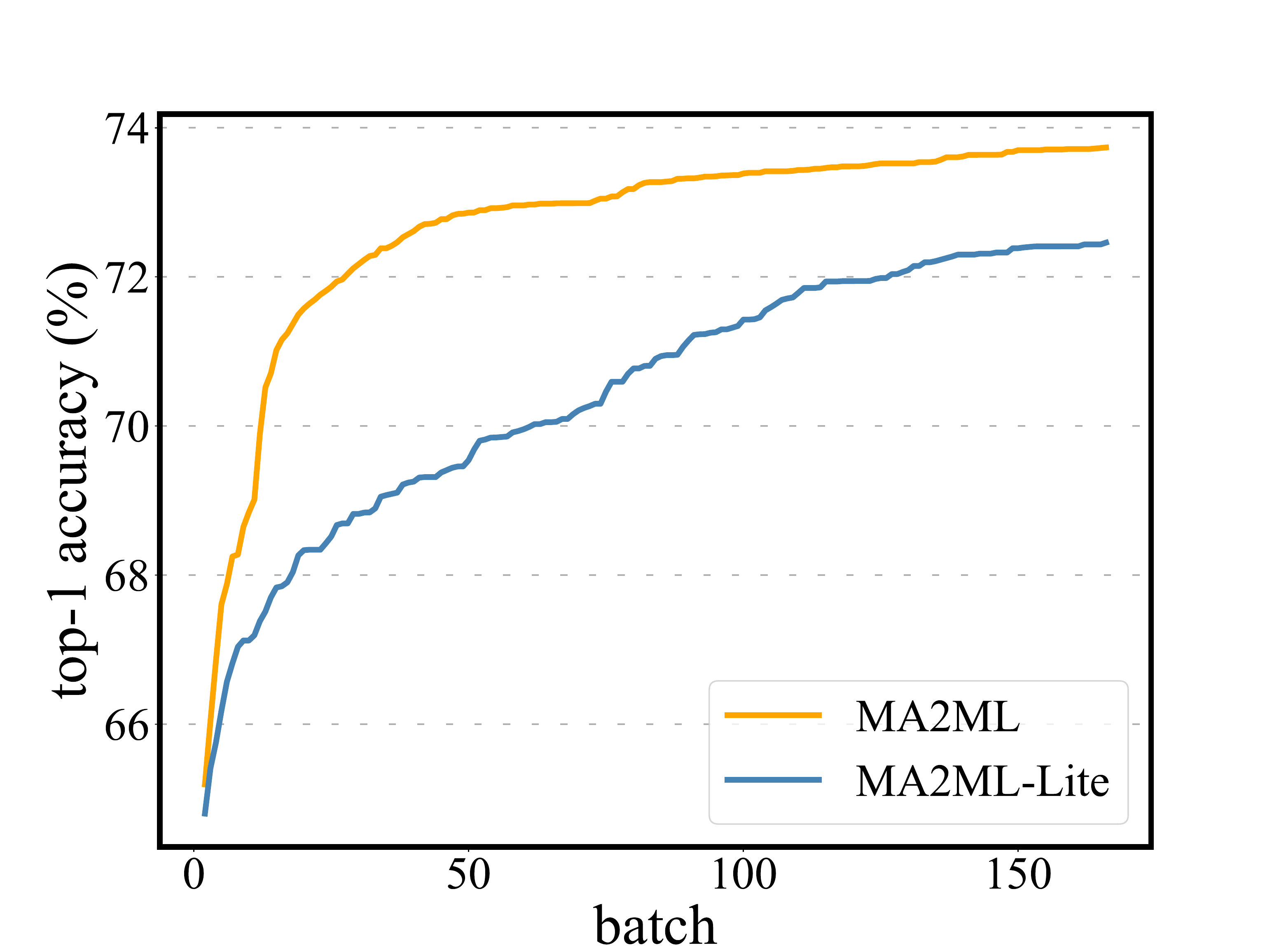}
    \includegraphics[width=0.23\textwidth]{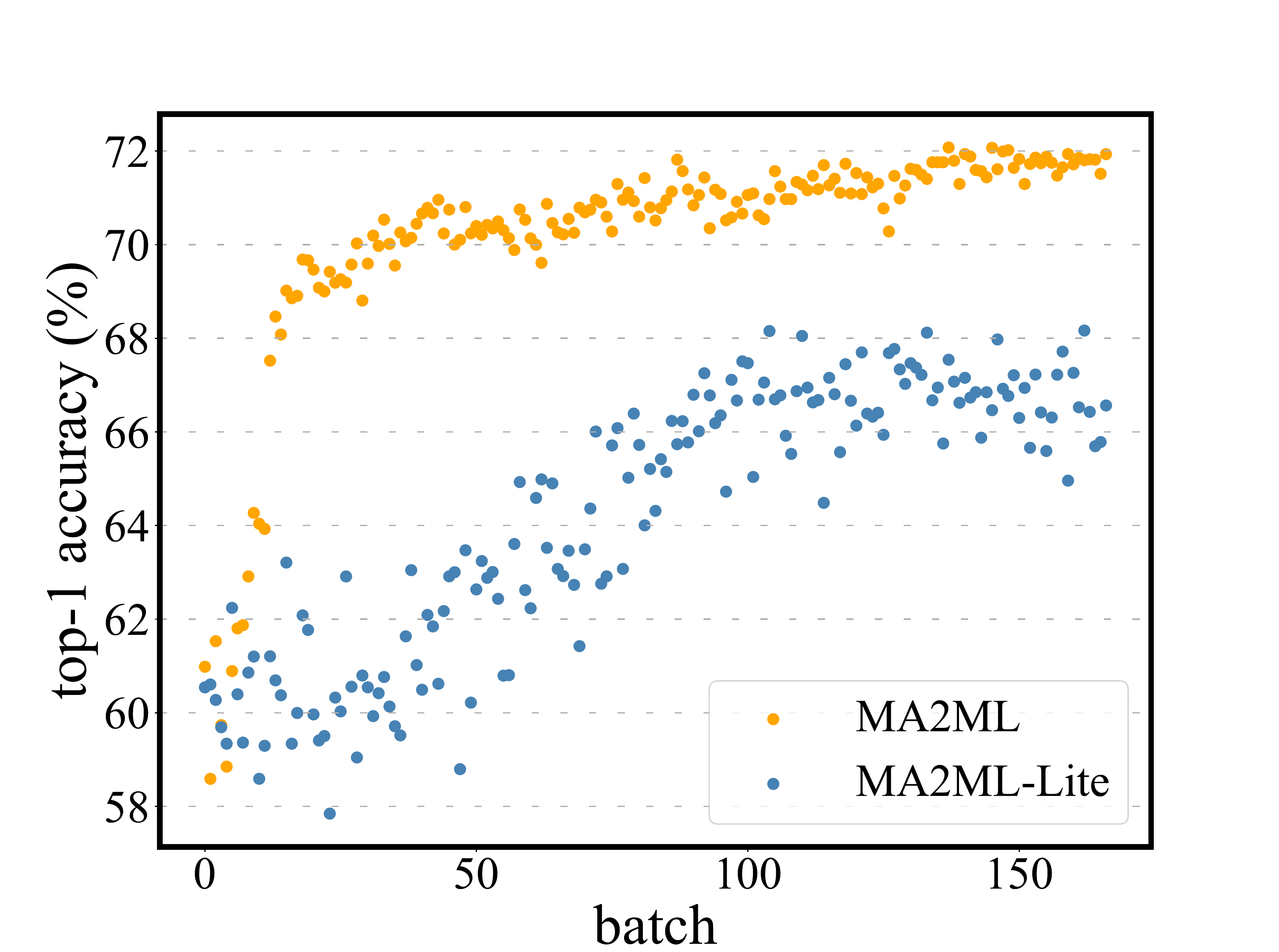}
    %\vspace{-3mm}
    \caption{Learning patterns of MA2ML and MA2ML-Lite on CIFAR-100. Left: average accuracy curves of top-30 pipelines in terms of batch numbers. Right: the scatter plot for the average accuracy of pipelines in each batch.}
    \label{fig: cifar100}
%\vspace{-4mm}
\end{figure}

Figure \ref{fig: cifar100} presents the average reward curves of top-30 pipelines and the scatter plot of the average reward in each batch of MA2ML and MA2ML-Lite on CIFAR-100. The large lead of MA2ML over MA2ML-Lite validates the effectiveness of MA2ML.

\section{Experiment Details}

\noindent\textbf{Detailed search space.} 
In Section \ref{cifar_exp}, we consider 15 augment operations, including ShearX/Y, TranslateX/Y, Rotate, AutoContrast, Invert, Equalize, Solarize, Posterize, Contrast, Color, Brightness, Sharpness, and Cutout. On CIFAR-10 and CIFAR-100, we adopt the same search space of NASNet \cite{zoph2018learning} for NAS, and use [0.16, 0.08, 0.04, 0.02] and [0.0008, 0.0004, 0.0002, 0.0001] as the search space of warmup learning rate and weight decay for HPO, respectively. On ImageNet-200, we utilize the same search space of FBNetV3 \cite{dai2021fbnetv3} to form NAS and HPO.

\vspace{1.5mm}
\noindent\textbf{RL setting.} 
In MA2ML and MA2ML-Lite, for each module we use a one-layer LSTM controller with 100 hidden units to generate action distributions. On CIFAR-10, CIFAR-100 and ImageNet-200 datasets, we set $\eta_\theta=0.0004$ and $\eta_\phi=0.005$ for the learning rate of policy parameter $\theta$ in \eqref{eq: actor_update} and critic parameter $\phi$ in \eqref{eq: critic_update}, respectively. We set $\tau=0.004$ on CIFAR and $\tau=0.012$ on ImageNet-200 as the moving average parameter for the update of target policy parameter $\bar{\theta}$ in \eqref{eq: target_policy_update}. We set $\lambda=0.2$ as the coefficient of KL divergence regularization.

\vspace{1.5mm}
\noindent\textbf{Training details.} 
For the short-term training on CIFAR-10 and CIFAR-100 in the search, the batch size is set to 64 per GPU. Two GPUs are used to train a model for 12 epochs with 1 epoch warm-up. For the retraining, the batch size is set to 64 per GPU. Eight GPUs are used to train a model for 600 epochs with 5 epochs warm-up. The learning rate of the model is 4 times larger than that of the short-term training. For the short-term training on ImageNet-200, the batch size is set to 128 per GPU. Eight GPUs are used to train a model for 100 epochs with 5 epochs warmup. For retraining on ImageNet, the batch size is set as 256 per GPU. Sixteen GPUs are used to train a model for 400 epochs with 5 epochs warmup. The learning rate of the model is 4 times larger than that of the short-term training.

\vspace{1.5mm}
\noindent\textbf{Detailed results.} 
On CIFAR-10 and CIFAR-100, the best searched pipeline is trained for 5 times. On CIFAR-10, MA2ML achieves 97.72\%, 97.84\%, 97.85\%, 97.75\%, and 97.7\%,  while MA2ML-Lite achieves 97.66\%, 97.59\%, 97.85\%, 97.6\%8, 97.73\% top-1 accuracy. On CIFAR-100, MA2ML achieves 84.83\%, 85.14\%, 85.15\%, 85.14\%, and 85.15\%, while MA2ML-Lite achieves 84.68\%, 84.92\%, 84.94\%, 84.73\%, 84.75\% top-1 accuracy. On ImageNet, the searched pipeline is trained for a one time. MA2ML-D/E/F achieve $80.5\%/80.7\%/81.1\%$ top-1 accuracy under 795M/890M/973M FLOPs.

\vspace{1.5mm}
\noindent\textbf{Search cost.}
Tesla V100s are used for experiments. On CIFAR-10/100 the search cost of MA2ML is 300 GPU days. On ImageNet the search cost is 2000 GPU days. For comparison, NASNet spends 2000 GPU days for search on CIFAR-10 with the same search space, which is much larger than MA2ML as the off-policy learning of MA2ML significantly improves sampling efficiency.

\end{document}